\documentclass[journal,twoside,web]{ieeecolor2}
\usepackage{generic}
\usepackage{cite}
\usepackage{amsmath,amssymb,amsfonts}
\usepackage{algorithmic}
\usepackage{graphicx}
\usepackage{textcomp}
\usepackage{bbm} 
\usepackage{multirow}
\usepackage{hyperref} 
\usepackage{tablefootnote}
\usepackage{threeparttable}
\usepackage{makecell}
\def\BibTeX{{\rm B\kern-.05em{\sc i\kern-.025em b}\kern-.08em
    T\kern-.1667em\lower.7ex\hbox{E}\kern-.125emX}}
\begin{document}
\title{Self-supervised Deep Unrolled Model with Implicit Neural Representation Regularization for Accelerating MRI Reconstruction}
\author{Jingran Xu, Yuanyuan Liu, Yuanbiao Yang, Zhuo-Xu Cui, Jing Cheng, Qingyong Zhu, Nannan Zhang, Yihang Zhou, Dong Liang,\IEEEmembership{Senior Member, IEEE} and Yanjie Zhu, \IEEEmembership{Senior Member, IEEE}
\thanks{This study was supported in part by the National Key R\&D Program of China nos. 2023YFA1011403, National Natural Science Foundation of China under grant nos. 62322119, 62201561, 62531024, 62206273.
(Corresponding author: Yanjie Zhu)}
\thanks{Jingran Xu, Yuanyuan Liu, Yuanbiao Yang, Jing Cheng, Nannan Zhang, and Yanjie Zhu are with Paul C. Lauterbur Research Center for Biomedical Imaging, Shenzhen Institutes of Advanced Technology, Chinese Academy of Sciences, Shenzhen, Guangdong, China(e-mail: $\left \{\text{jr.xu; liuyy; yb.yang; jing.cheng; nn.zhang2; yj.zhu}\right \}$@siat.ac.cn).}
\thanks{Zhuo-Xu Cui, Qingyong Zhu,Yihang Zhou, and Dong Liang are with Research Center for Medical AI, Shenzhen Institutes of Advanced Technology, Chinese Academy of Sciences, Shenzhen, Guangdong, China(e-mail: $\left \{\text{zx.cui; qy.zhu; yh.zhou2; dong.liang}\right\}$@siat.ac.cn)}
\thanks{Jingran Xu and Yuanyuan Liu contributed equally to this study.}
}

\maketitle

\begin{abstract}
Magnetic resonance imaging (MRI) is a vital clinical diagnostic tool, yet its application is limited by prolonged scan times. Accelerating MRI reconstruction addresses this issue by reconstructing high-fidelity MR images from undersampled $k$-space measurements. In recent years, deep learning-based methods have demonstrated remarkable progress. However, most methods rely on supervised learning, which requires large amounts of fully-sampled training data that are difficult to obtain. This paper proposes a novel zero-shot self-supervised reconstruction method named UnrollINR, which enables scan-specific MRI reconstruction without external training data. UnrollINR adopts a physics-guided unrolled reconstruction architecture and introduces implicit neural representation (INR) as a regularization prior to effectively constrain the solution space. This method overcomes the local bias limitation of CNNs in traditional deep unrolled methods and avoids the instability associated with relying solely on INR's implicit regularization in highly ill-posed scenarios. Consequently, UnrollINR significantly improves MRI reconstruction performance under high acceleration rates. Experimental results show that even at a high acceleration rate of 10, UnrollINR achieves superior reconstruction performance compared to supervised and self-supervised learning methods, validating its effectiveness and superiority.
\end{abstract}


\section{Introduction}
\label{sec:introduction}
\IEEEPARstart{M}{agnetic} resonance imaging (MRI) is a non-invasive, radiation-free technique that has become indispensable in clinical practice. However, its inherently long acquisition time remains a major technical limitation. Therefore, fast imaging techniques based on $k$-space undersampling have attracted considerable attention\cite{huang2024data}. 
Parallel imaging (PI) and compressed sensing (CS) are the most widely used traditional approaches. PI exploits the redundancy among multi-channel receiver coils to reconstruct images from undersampled data, whereas CS leverages the compressibility of images through pseudo-random sparse sampling combined with sparsity or low-rank constraints~\cite{lustig2010spirit, lustig2007sparse}. Despite their success, both techniques face challenges in achieving high acceleration rates\cite{sandino2020compressed}.

In recent years, deep learning has achieved remarkable progress in accelerating MRI reconstruction. 
Early data-driven approaches learn an end-to-end mapping from undersampled $k$-space or artifact-corrupted images to fully sampled or artifact-free images using neural networks~\cite{knoll2020deep, zhu2018image}. Although straightforward, these methods lack physical interpretability and typically require large-scale, fully-sampled datasets to characterize such mappings, which is often impractical in clinical settings. 
To overcome these limitations, physics-guided reconstruction frameworks have been developed, combining model-based data fidelity with learned regularization terms to explicitly incorporate MRI physics~\cite{hammernik2023physics}.
A representative paradigm is the unrolled framework, which unrolls the traditional iterative reconstruction algorithms into deep networks. Within this framework, each iteration is expanded into a network block by learning regularization terms, penalty parameters, or update rules~\cite{monga2021algorithm}.
Typical algorithms include ADMM-Net~\cite{sun2016deep}, ISTA-Net~\cite{zhang2018ista}, MoDL~\cite{aggarwal2018modl}, and VarNet~\cite{sriram2020end}. 
Compared with data-driven approaches, deep unrolled methods significantly enhance reconstruction performance and model interpretability.

Although unrolled methods have shown promising results, several challenges remain to be addressed. First, most unrolled methods rely on supervised learning, which requires fully sampled training datasets and exhibits limited generalization capability to out-of-distribution data~\cite{montalt2021machine, antun2020instabilities}. 
Zero-shot unrolled method has been proposed to mitigate this issue by learning the reconstruction prior exclusively from the undersampled measurement itself but its performance still lags behind supervised counterparts~\cite{yaman2022zero}. 
Second, unrolled methods typically employ convolutional neural networks (CNNs) as the regularization backbone, whose limited receptive field restricts their ability to capture global context. Expanding the receptive field requires stacking deeper networks, which increases computational cost and makes performance highly architecture-dependent. Finally, the optimal number of unrolled iterations remains uncertain and is often empirically chosen, which may lead to suboptimal convergence or unnecessary computation.

To address the aforementioned challenges, this study proposes UnrollINR, a novel zero-shot self-supervised reconstruction framework that employs implicit neural representations (INRs) to model the regularization terms within unrolled reconstruction methods. 
The framework enforces data consistency through a forward physical model while embedding INR-output as an effective prior into the unrolled optimization process.
Specifically, INR models image intensities as a continuous function of spatial coordinates via a network, typically a multi-layer perceptron (MLP), providing a continuous image representation~\cite{feng2025spatiotemporal}. 
The MLP’s global connectivity enables it to naturally capture global and long-range dependencies, effectively alleviating the locality bias of CNNs.
By integrating INR as a regularization term yields a compact unrolled architecture that eliminates the need for deep stacked network structures.
This design achieves robust regularization and enhances fine detail recovery utilizing only a shallow MLP within a single unrolled iteration.
Furthermore, this integration mitigates the inherent instability typically encountered when relying solely on INR's regularization capability in highly ill-posed scenarios.
We evaluated the performance of UnrollINR on two public datasets and one private dataset. Both retrospective and prospective experimental results demonstrate that our method outperforms comparative supervised and self-supervised learning methods. Even at high acceleration rates, it achieves robust reconstruction performance using only undersampled $k$-space data, fully validating its superiority. Moreover, the lightweight network significantly reduces the training cost.

The main contributions of this work are summarized as follows:
\begin{enumerate}
\item We propose a novel zero-shot self-supervised unrolled reconstruction framework for accelerating MRI reconstruction, which achieves superior performance using only undersampled $k$-space data from a single subject.
\item We introduce INR as an explicit regularization prior within the unrolled reconstruction framework, enabling continuous spatial modeling and effective global-context regularization that alleviates the locality bias of CNN.
\item We develop a compact architecture that integrates data-consistency enforcement with INR-based regularization in a single unrolled iteration, achieving stable and high-quality reconstructions at high acceleration rates while substantially reducing computational burden.
\end{enumerate}

\section{Related work}
\subsection{Deep Unrolled Method}
Physics-guided deep unrolled iterative methods solve the inverse problem of MRI reconstruction by unrolling traditional iterative algorithms into network structures~\cite{yaman2020self}. These methods incorporate prior knowledge through learnable regularization while enforcing data fidelity constraints derived from the physical imaging model, thereby enhancing both reconstruction accuracy and interpretability. The primary differences among various deep unrolled methods lies in the neural network architecture employed for the regularization term and the solving algorithm adopted for the data consistency term~\cite{yaman2022zero}.

Currently, a wide range of deep unrolled methods have been developed based on various optimization schemes, including the alternating direction method of multipliers (ADMM)~\cite{chan2016plug}, the iterative shrinkage-thresholding algorithm (ISTA)~\cite{beck2009fast}, the proximal gradient descent method (PGD)~\cite{mardani2018neural}, and the variable splitting with quadratic penalty~\cite{afonso2010fast}. Among these, ADMM-Net pioneered the application of the ADMM framework to CS-MRI~\cite{sun2016deep}. ISTA-Net unrolled the ISTA algorithm into multiple network stages, each employing learnable modules for data consistency and regularization~\cite{zhang2018ista}. MoDL adopted a variable splitting strategy with a CNN-based regularizer, providing a systematic framework for deep architectures in inverse problems~\cite{aggarwal2018modl}. However, all the aforementioned methods rely on supervised learning, which requires large amounts of fully-sampled training data for satisfactory performance. 
To alleviate this limitation, self-supervised and zero-shot frameworks such as SSDU~\cite{yaman2020self} and ZS-SSL~\cite{yaman2022zero} have been proposed.
SSDU enables unrolled iterative reconstruction using only undersampled data, while ZS-SSL achieves subject-specific reconstruction by training on measurements from a single undersampled scan.
Nevertheless, these approaches still face challenges in reconstruction quality and computational burden, as multiple iterative unrolling steps are required during training, leading to high memory and time costs.

\subsection{Implicit Neural Representation}
In recent years, INR has garnered significant attention in the field of computer vision. The core idea is to use a neural network, typically a MLP, to model an image or volumetric data as a continuous function of spatial coordinates~\cite{sitzmann2020implicit, park2019deepsdf}. 
Specifically, the network takes spatial coordinates as input and outputs the corresponding image intensity values. Once the network training converges, the continuous representation of the image is implicitly stored within its weights.
The underlying network can be formulated as:
\begin{equation}
 f_{\theta}: v=(v_x,v_y) \in \mathbb{R}^{2} \rightarrow I \in \mathbb{C},
\label{eq8}
\end{equation}
where $v$ represents the coordinate and $I$ denotes the corresponding image intensity. 

Meanwhile, related studies have shown that INR networks exhibit significant limitations in representing high-frequency details when processing raw input coordinates directly~\cite{rahaman2019spectral}. 
To address this issue, coordinate encoding functions have been proposed that map input coordinates into a higher-dimensional space, thereby enhancing the model's capacity to capture high-frequency information. 
These encoding functions can be primarily categorized as fixed encodings~\cite{liu2022recovery} and learnable encodings~\cite{zhu2024disorder}. 
Fixed encodings employ predefined transformation rules, such as positional
encoding~\cite{mildenhall2021nerf}, Fourier feature encoding~\cite{tancik2020fourier}.
In contrast, learnable encodings introduce learnable parameters and efficient sparse data structures, achieving superior convergence performance and reconstruction accuracy~\cite{takikawa2021neural, liu2020neural}. 
When the encoding function is adopted, \eqref{eq8} can be rewritten as: 
\begin{equation}
 f_{\theta}: \phi(v) \rightarrow I \in \mathbb{C},
\label{eq9}
\end{equation}
where $\phi(\cdot)$ represents the coordinate encoding function.

\subsection{INR-based MR reconstruction}
Currently, INR has demonstrated considerable potential in accelerated MRI reconstruction. For instance, NeRP proposes INR learning with prior embedding to reconstruct images from radially undersampled $k$-space data~\cite{shen2022nerp}. IMJENSE integrates INR with PI to achieve joint estimation of coil sensitivity maps and images~\cite{feng2023imjense}. 
Furthermore, INR-based approaches have been applied to dynamic and quantitative MRI. 
In dynamic MRI, methods such as CineJENSE~\cite{al2023cinejense}, FMLP~\cite{kunz2024implicit}, and IMJ-PLUS~\cite{shen2025imj} model the image sequence as a continuous neural function across spatial and temporal domains. 
In quantitative MRI, algorithms including INR-QSM~\cite{zhang2024subject}, PhysINR~\cite{liu2025physics} and SUMMIT~\cite{lao2025coordinate} recover tissue-specific physical parameters directly from undersampled data.
Despite these advances, most INR-based methods predominantly rely on the implicit regularization capability of the network itself. Under high acceleration rates, these methods often exhibit instability, which limits their reconstruction performance.

\section{Methodology}
\subsection{Overall Framework}
\begin{figure*}[!t]
\centerline{\includegraphics[width=12cm]{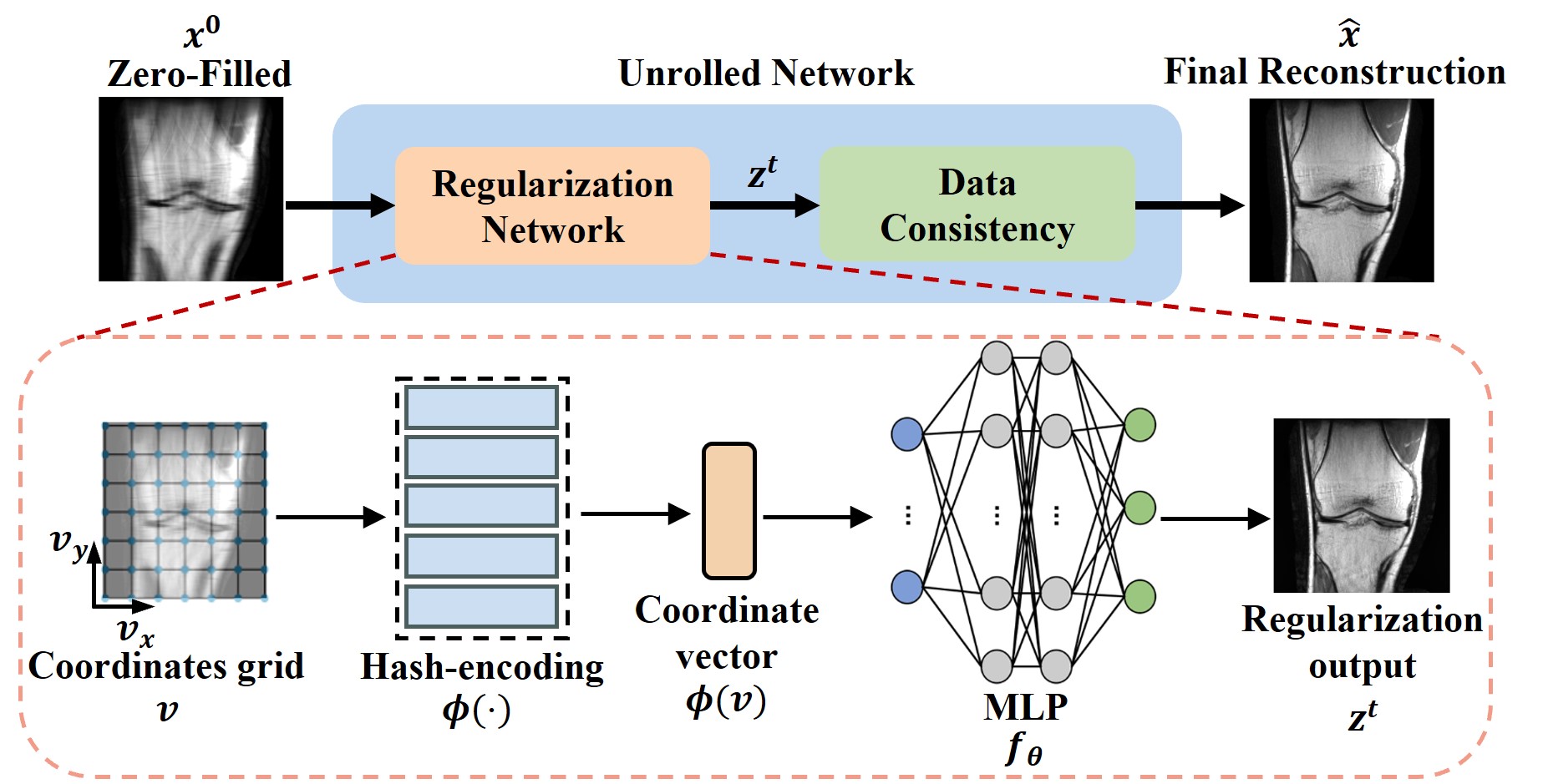}}
\caption{Overview of the proposed UnrollINR framework with an unrolled iterative architecture incorporating an INR-based network as the regularization term.}
\label{fig1}
\end{figure*}

The overall framework is illustrated in Fig. \ref{fig1}. UnrollINR adopts a physics-guided unrolled architecture, in which the reconstruction network consists of a regularization module and a data consistency (DC) module. The regularization module is implemented via an INR-based neural network that captures the inherent prior information of the data, while the DC module is solved using the conjugate gradient (CG) method, to ensure fidelity to the physical acquisition model. The network input, $x^{0}$, is initialized by applying an inverse Fourier transform to the zero-filled undersampled $k$-space measurement $y$, i.e., $x^{0}=E^H y$. The network output, $ \widehat{x}$, represents the final high-fidelity reconstructed image.

\subsection{Reconstruction Model for Accelerated MRI}
In multi-channel MRI, multiple receiver coils are utilized to sample frequency-domain raw data, also known as $k$-space. To accelerate the acquisition process, $k$-space measurements are often undersampled according to a predefined sampling mask. Accordingly, the forward model of the multi-coil MRI acquisition can be expressed as:
\begin{equation}
y_i=MFC_i x+n_i,
\label{eq1}
\end{equation}
where $x$ denotes the image to be reconstructed, $y_i$ is the measurement from the $i$th coil, $n_i$ is the corresponding noise, $M$ is the undersampling mask, $F$ represents the Fourier transform operator, and $C_i$ is the sensitivity map matrix. Consequently, the MR acquisition model can be compactly formulated across the coil dimension as:
\begin{equation}
y=Ex+n,
\label{eq2}
\end{equation}
where $E$ represents the forward encoding operator, constructed by concatenating $MFC_i$ across the coil dimension. 
The reconstruction of $x$ can be formulated as the optimization problem:
\begin{equation}
\arg \min_{x} \left\| y-Ex \right\|_{2}^{2}
+ \mu \mathcal{R}({x}),
\label{eq3}
\end{equation}
where the first term enforces data fidelity with the acquired measurement $y$, $\mathcal{R}({x})$ denotes the regularization term that incorporates prior knowledge into the reconstruction, and $\mu$ is a regularization parameter balancing these two terms.

\subsection{Physics-guided Unrolled Iterative Architecture}
In this study, the variable splitting strategy with quadratic penalty is employed to reformulate \eqref{eq3} as: 
\begin{equation}
\arg \min_{x,z} \left\| y-Ex \right\|_{2}^{2} + \lambda \left\| x-z \right\|_{2}^{2}
+ \mathcal{R}({z}),
\label{eq4}
\end{equation}
where $z$ is an auxiliary variable, $\lambda$ is a regularization parameter. Hence, the optimization problem in \eqref{eq4} can thus be decomposed into the following two sub-problems as: 
\begin{equation}
z^{t}= \arg \min_{z} \lambda \left\| x^{t-1} -z\right\|_{2}^{2} + 
\mathcal{R}({z}),
\label{eq5}
\end{equation}
\begin{equation}
x^{t}=\arg \min_{x} \left\| y-Ex \right\|_{2}^{2} + 
\lambda \left\| x-z^{t} \right\|_{2}^{2},
\label{eq6}
\end{equation}
where $x^{0}$ is the initial zero-filled reconstruction derived from the measurement $y$, with $x^{t}$ and $z^{t}$ being the estimated image and an intermediate variable at the $t$th iteration, respectively. These two problems are addressed alternatively within each iteration. Specifically, the regularization subproblem in \eqref{eq5} is implicitly solved by an INR-based neural network, while the DC subproblem in \eqref{eq6} is solved using the normal equation:
\begin{equation}
x^{t}= (E^HE + \lambda I)^{-1} (E^H y + \lambda z^{t}),
\label{eq7}
\end{equation}
where $I$ denotes the identity operator and $(\cdot)^H$ denotes the conjugate transpose operator. In practical MRI, the acquired data are often multi-channel, and $(E^HE + \lambda I)^{-1}$ in \eqref{eq7} is not analytically invertible. Therefore, \eqref{eq7} is often solved using iterative numerical methods such as conjugate gradient (CG) and gradient descent algorithms.

The DC term is optimized using the CG algorithm. Although CG inherently requires multiple iterative steps to accurately enforce data fidelity constraints, it contains no trainable parameters. This enables backpropagation without storing intermediate CG results, allowing extensive CG iterations per training iteration with almost no increase in memory usage.

\subsection{the Proposed Deep Unrolled Network}
The proposed method, UnrollINR, incorporates an INR-based neural network as the regularization term to exploit image priors within the MRI reconstruction framework. This INR-based regularization term establishes a mapping between MRI spatial coordinates and their corresponding image intensities through a learnable continuous mapping function. As shown in Fig. \ref{fig1}, the continuous mapping function $f_{\theta}$ is approximated by an MLP. To enhance the model's capacity for fitting high-frequency details, the input coordinates $v$ are first mapped into a higher-dimensional feature space using a learnable coordinate encoding function $\phi(\cdot)$ before being fed into the MLP. The process is represented as: 
\begin{equation}
    z^{t}=  f_{\theta} (\phi(v)),
   \label{eq10}
\end{equation}
where $z^{t}$ represents the intermediate variable generated by the $t$th iteration.

This study employs multi-resolution hash encoding as the coordinate encoding function $\phi(\cdot)$~\cite{muller2022instant}. The combination of hash encoding with an MLP enables the use of a shallower MLP to achieve faster convergence and superior detail reconstruction performance. Specifically, the multi-resolution hash encoding organizes trainable parameters into $L$ distinct resolution levels, each corresponding to an independent hash table that stores learnable feature vectors. Each level contains $T$ feature vectors of dimension $F$, resulting in $T\times F$ learnable parameters per level. The resolutions of the hash grids are arranged in a geometric progression: $N_{min}$, $b \times N_{min}$, $\ldots$, $b^{(L-1)} \times N_{min}$, where $N_{min}$ and $b$ denote the initial term and common ratio of the progression, respectively. Coarser grid nodes cover larger coordinate regions, thereby helping to preserve non-local continuity of the signal, while finer grid nodes correspond to local coordinates, enabling the capture of richer high-frequency details. For any given input coordinate $v$, the encoding function retrieves the feature vectors of the corresponding voxel vertices at each resolution level and performs linear interpolation based on the coordinate's relative position within the voxel. Finally, the interpolated $F$-dimensional feature vectors from all $L$ levels are concatenated to form the final coordinate encoding vector $\phi(v)$ with an output dimension of $L\times F$.
\subsection{Loss Function}
The loss function is composed of a data fidelity term and a total variation-based regularization term, expressed as:
\begin{equation}
    L_{total}= L_{DC}+\lambda_{s}L_{TV},
   \label{eq11}
\end{equation}
where $\lambda_{s}$ is the penalty parameter that balances the two terms.

Data consistency with the original undersampled $k$-space measurement $y$ is ensured by minimizing the data fidelity term $L_{DC}$, which is defined using a normalized $l_1-l_2$ loss as:
\begin{equation}
    L_{DC}= \frac{\|y-\widehat{y}\|_{2}}{\|y\|_{2}}+\frac{\|y-\widehat{y}\|_{1}}{\|y\|_{1}},
   \label{eq12}
\end{equation}
where $\widehat{y}$ is obtained by transforming the final reconstructed output image $\widehat{x}$ of the network into $k$-space using the forward model $E$, that is, $\widehat{y}= E\widehat{x}$.

Minimizing the total variation (TV) loss $L_{TV}$ can enhance local spatial consistency, eliminate image noise, and preserve edges in the reconstructed image. This loss operates directly on the final reconstructed image $\widehat{x}$ and is formulated as:
\begin{equation}
    L_{TV}= \|G\widehat{x}\|_{1},
   \label{eq13}
\end{equation}
where $G$ represents the gradient operator.

\section{Experiments}
\subsection{Datasets}
Retrospective experiments utilized publicly available fully-sampled multi-coil knee and brain data from the fastMRI dataset~\cite{zbontar2018fastmri}. The knee data consisted of T1-weighted images acquired with 15 receiver coils. All knee images were center-cropped to a size of $368\times 368$. To train the supervised methods, 739 slices from 29 subjects were selected as the training dataset. The brain data consisted of T2-weighted images acquired with 20 receiver coils. All brain images were center-cropped to a size of $320\times 320$. Similarly, for training the supervised methods, 150 slices from 15 subjects were selected as the training dataset.
For retrospective undersampling, random undersampling masks were employed.
Table \ref{tab1} summarizes the datasets, acceleration rates R, actural undersampling rates u\_rate, and ACS sizes. Note that R $\approx$ 1/u\_rate.
\begin{table}[htbp]
\centering
\caption{Datasets and the Corresponding acceleration rates R, undersampling rates u\_rate, and ACS sizes}
\label{tab1}
\renewcommand{\arraystretch}{1.2}
\begin{tabular}{c|ccc}
\hline
\textbf{Datasets}&  \textbf{R}&  \textbf{u\_rate}& \textbf{ACS size}\\    \hline
\multirow{3}{*}{\makecell{\textbf{FastMRI} \\ \textbf{knee dataset}}}&   6&   16.58\%&   18\\  \cline{2-4}                                        &     8&   12.50\%&   16\\   \cline{2-4} 
                              &      10&   9.78\%&   12\\    \hline
\multirow{3}{*}{\makecell{\textbf{FastMRI} \\ \textbf{brain dataset}}}&   6&   16.56\%&   18\\  \cline{2-4}                                        &     8&   12.50\%&   16\\   \cline{2-4} 
                              &      10&   10.00\%&   12\\    \hline
\end{tabular}
\end{table}
\par The prospectively undersampled data were acquired using a 3T scanner (TIM TRIO, Siemens, Erlangen, Germany) with a 12-channel head coil, employing a fast spin-echo sequence. The study was approved by the local institutional review board. The imaging parameters were as follows: TE/TR = 5.8 ms/4 s; echo train length = 8; space resolution = $0.7\times0.7$; matrix size = $384\times 384$.  Similarly, for training the supervised methods, an additional 112 fully-sampled slices from 7 subjects were used as the training dataset. For prospective undersampling, random mask with u\_rate = 12.5\% was employed. 

\subsection{Performance Evaluation}
The proposed method was compared with several state-of-the-art MRI reconstruction methods, including MoDL~\cite{aggarwal2018modl}, ZS-SSL~\cite{yaman2022zero}, IMJENSE~\cite{feng2023imjense}, ConvDecoder~\cite{darestani2021accelerated}, and L1-ESPIRiT~\cite{uecker2014espirit}.
MoDL is a supervised unrolled iterative method that incorporates a CNN-based regularizer~\cite{aggarwal2018modl}. 
ZS-SSL is a scan-specific self-supervised approach also based on an unrolled iterative framework with CNN regularization~\cite{yaman2022zero}.
IMJENSE introduces a scan-specific self-supervised method leveraging INR to learn a continuous functional representation of MR images directly from the undersampled $k$-space measurement~\cite{feng2023imjense}.
ConvDecoder addresses fast MRI reconstruction using a deep image prior (DIP)-based scan-specific approach~\cite{darestani2021accelerated}.
L1-ESPIRiT is a traditional parallel MRI reconstruction method employing total variation as the regularization term, implemented using the BART toolkit~\cite{uecker2014espirit, blumenthal2022bart}.
All compared methods were executed according to the typical setting mentioned by the authors.

For quantitative assessment, the peak signal-to-noise ratio (PSNR), structural similarity index (SSIM), and normalized root mean squared error (NRMSE) were employed as evaluation metrics. 

\subsection{Implementation Details}
The multi-resolution hash encoding and MLP within the INR-based regularization term were implemented using the tiny CUDA neural networks~\cite{tiny-cuda-nn} library. The MLP contained two hidden layers with 64 neurons each. The data consistency term was solved using the CG method, with the number of internal iterations for this module set to 20. For all datasets used in the experiments, coil sensitivity maps were estimated from the central region of the fully sampled $k$-space using ESPIRiT with default parameters~\cite{uecker2014espirit}.

Regarding the selection of hyperparameters, the proposed UnrollINR involves two hyperparameters: the regularization parameter $\lambda$ in the unrolled framework and the penalty parameter $\lambda_{s}$ in the loss function.
These two hyperparameters were set as learnable parameters. They were initialized with specific values at the beginning of the experiment and were automatically updated during the training process. The selection of these initial values will be analyzed in subsequent experiments.
For the retrospective experiments, the initial values were set as $\lambda = 0.01$ and $\lambda_{s} = 0.5$.
For the prospective experiments, the initial values were set as $\lambda = 0.05$ and $\lambda_{s} = 2.0$.

The proposed method was implemented in PyTorch 1.10 using Python 3.9, and all experiments were conducted on a workstation equipped with an NVIDIA A100 GPU (80 GB).

\subsection{Ablation Study}
We conducted three ablation studies to evaluate the impact of key components in the proposed UnrollINR method on the reconstruction performance. 
First, the effectiveness of the unrolled network was validated. This unrolled network consists of a regularization term and a data consistency term. We evaluated their impact on reconstruction performance by removing each term sequentially.
Second, the influence of the coordinate encoding method within the regularization term was examined. We compared the default Instant-NGP~\cite{muller2022instant} encoding with DINER~\cite{zhu2024disorder}, both of which are learnable encoding approaches widely used in various tasks.
Finally, the contribution of the loss function was investigated. By removing the TV regularization term, its effect on the overall reconstruction performance was assessed.

\section{Results}
\subsection{Retrospective Reconstruction}
\begin{figure*}[!t]
\centerline{\includegraphics[width=15cm]{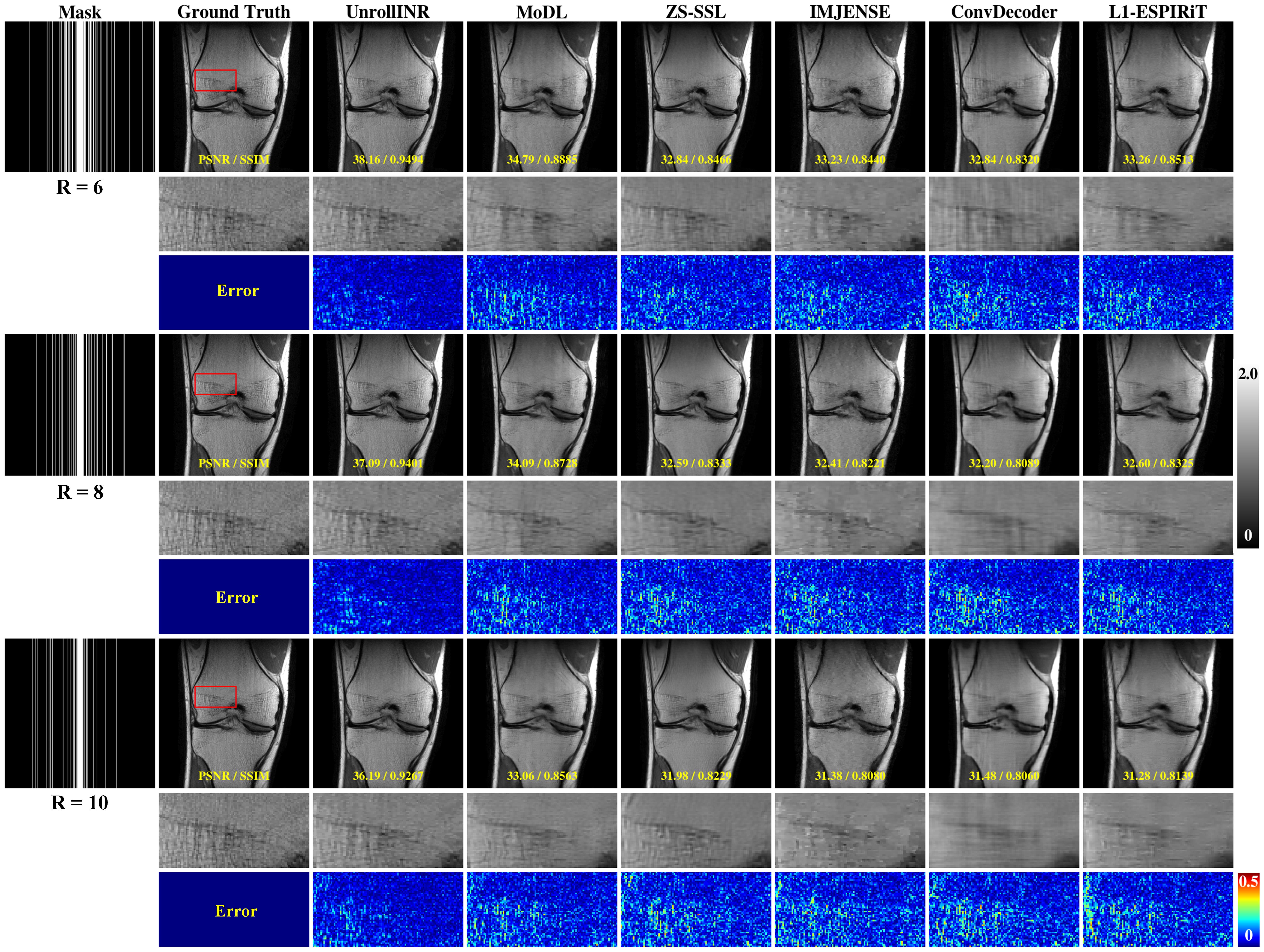}}
\caption{Comparative results of different methods on the fastMRI knee dataset under various acceleration rates R. The quantitative metrics PSNR and SSIM are indicated at the bottom of each reconstructed image. Local magnified views of the reconstructed images and absolute error maps are provided. The undersampling masks used for the corresponding acceleration rates R are displayed on the far left of the figure.}
\label{fig2}
\end{figure*}

Fig. \ref{fig2} presents the retrospective reconstruction results of all comparative methods on a randomly selected knee slice under different acceleration rates. The PSNR and SSIM values are annotated beneath each reconstruction. Local magnified views alongside absolute error maps are provided for detailed comparison. The undersampling masks are shown at the far left of the figure. Based on the quantitative metrics, the proposed UnrollINR achieves the best reconstruction performance across all three acceleration rates, significantly outperforming other comparative methods. In terms of visual results, the unsupervised methods ConvDecoder and L1-ESPIRiT exhibit noticeable artifacts and overly smooth outcomes. The images reconstructed using IMJENSE show apparent blurring and loss of fine details. The unrolled method ZS-SSL effectively removes undersampling artifacts but requires further improvement in detail recovery. The supervised method MoDL also exhibits noticeable artifacts. In contrast, the proposed UnrollINR effectively eliminates undersampling artifacts and achieves the recovery of fine details.

\begin{figure*}[!t]
\centerline{\includegraphics[width=15cm]{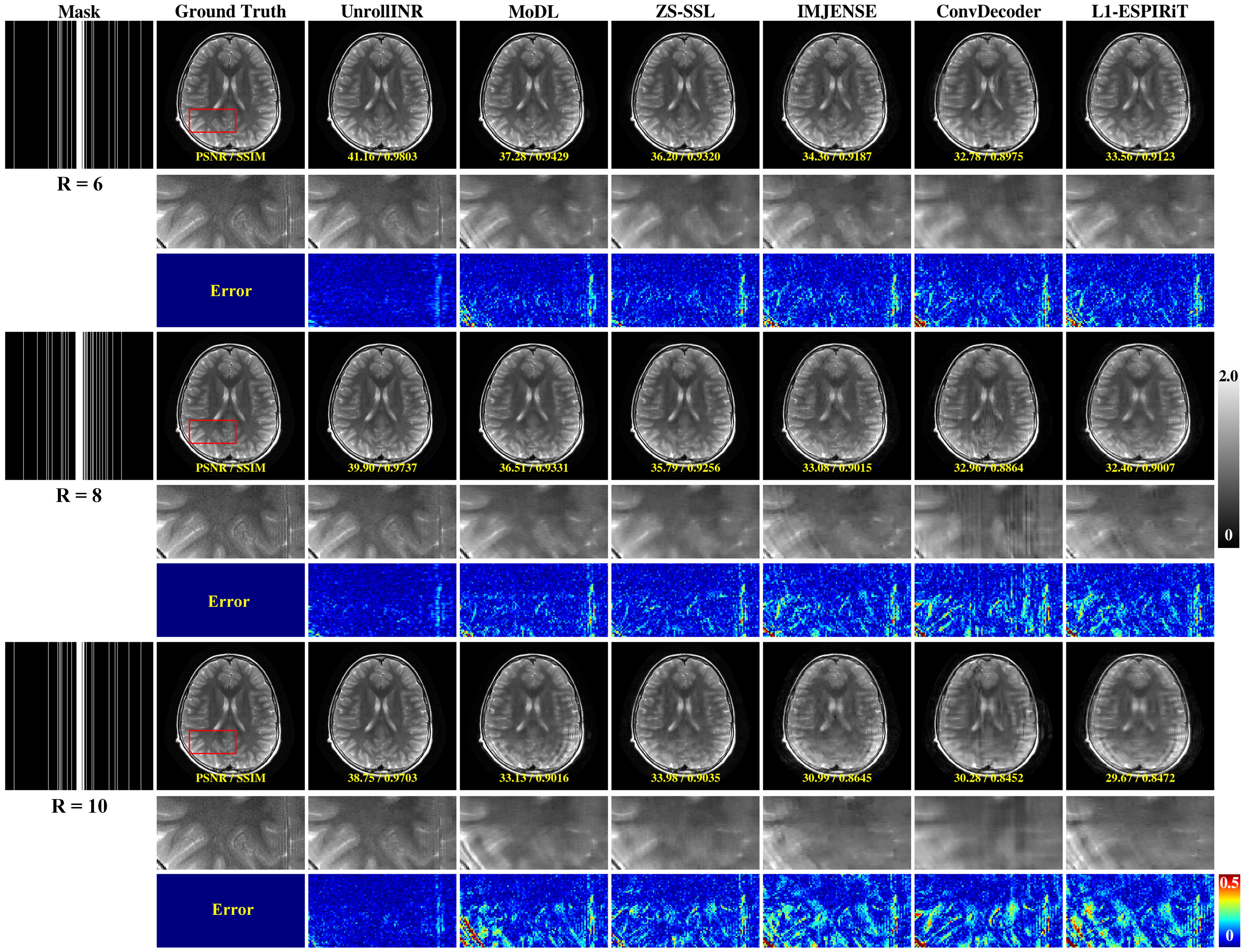}}
\caption{Comparative results of different methods on the fastMRI brain dataset under various acceleration rates R. The quantitative metrics PSNR and SSIM are indicated at the bottom of each reconstructed image. Local magnified views of the reconstructed images and absolute error maps are provided. The undersampling masks used for the corresponding acceleration rates R are displayed on the far left of the figure.}
\label{fig3}
\end{figure*}

Fig. \ref{fig3} presents the retrospective reconstruction results of all comparative methods on a randomly selected brain slice under different acceleration rates. According to the quantitative evaluation metrics of the reconstructed images, the proposed UnrollINR again achieves the best reconstruction performance. Visually, UnrollINR also demonstrates effective artifact removal and fine detail recovery. The supervised method MoDL shows noticeable residual artifacts at high acceleration rates. The unrolled method ZS-SSL loses fine details in its reconstructed images. Other comparative unsupervised methods exhibit significant noise and residual artifacts.

To further evaluate the proposed method, retrospective experiments were conducted on 30 randomly selected slices from the fastMRI knee dataset under acceleration rates of R = 6, 8, and 10. Table \ref{tab2} summarizes the results, showing the mean and standard deviation of PSNR, SSIM, and NRMSE over 30 slices. The proposed UnrollINR demonstrated the best performance among all comparative methods, with the highest PSNR and SSIM values and the lowest NRMSE value across all three acceleration rates. These results are consistent with the aforementioned visual results, further confirming the superiority of UnrollINR.
\begin{table*}[htbp]
\centering
\caption{Reconstruction results for all comparative methods on the fastMRI knee dataset under different acceleration rates R. The values represent the mean and standard deviation of the quantitative evaluation metrics (PSNR, SSIM, and NRMSE) across 30 slices, with the optimal results highlighted in bold.}
\label{tab2}
\renewcommand{\arraystretch}{1.2}
\begin{tabular}{cc|cccccc}
\hline
\textbf{R}& \textbf{Metrics}&  \textbf{UnrollINR}&  \textbf{MoDL}&  \textbf{ZS-SSL}&  \textbf{IMJENSE}&  \textbf{ConvDecoder}&  \textbf{L1-ESPIRiT}\\    \hline
\multirow{2}{*}{6}&    PSNR&   \textbf{39.39$\pm$1.63} &     36.14$\pm$1.49 &       $33.53\pm1.75$ &        34.57$\pm$1.72 &   33.57$\pm$1.33 &    34.18$\pm$1.20 \\

                    &    SSIM&   \textbf{0.9606$\pm$0.0082} &  0.9169$\pm$0.0170 &    0.8840$\pm$0.0236 &        0.8869$\pm$0.0247 &    0.8708$\pm$0.0235 &    0.8851$\pm$0.0200 \\    

                    &   NRMSE&   \textbf{0.0109$\pm$0.0019} &   0.0158$\pm$0.0026 &    0.0215$\pm$0.0044 &       0.0190$\pm$0.0034 &   0.0212$\pm$0.0031 &   0.0197$\pm$0.0026\\    \hline                   
                    
\multirow{2}{*}{8}&    PSNR&   \textbf{38.68$\pm$1.70} &     35.34$\pm$1.44 &       33.21$\pm$1.55 &        33.80$\pm$1.69 &   32.65$\pm$1.19 &     33.13$\pm$1.09 \\ 

                    &    SSIM&   \textbf{0.9545$\pm$0.0099} &  0.9047$\pm$0.0195 &    0.8723$\pm$0.0252 &  0.8706$\pm$0.0286 &     0.8507$\pm$0.0256 &    0.8663$\pm$0.0219 \\  

                    &   NRMSE&   \textbf{0.0119$\pm$0.0022} &   0.0173$\pm$0.0028 &    0.0222$\pm$0.0039 &       0.0208$\pm$0.0037 &   0.0235$\pm$0.0031 &   0.0222$\pm$0.0028\\    \hline
                    
\multirow{2}{*}{10}&   PSNR&   \textbf{37.72$\pm$1.66} &     33.47$\pm$1.19 &       31.71$\pm$1.45 &        32.54$\pm$1.53 &     31.08$\pm$1.29 &    31.47$\pm$1.03\\  

                     &   SSIM&   \textbf{0.9429$\pm$0.0119} &   0.8771$\pm$0.0213 &    0.8358$\pm$0.0357 &       0.8460$\pm$0.0309 &   0.8205$\pm$0.0280 &   0.8361$\pm$0.0225\\    

                     &  NRMSE&   \textbf{0.0132$\pm$0.0024} &   0.0214$\pm$0.0029 &    0.0263$\pm$0.0045 &       0.0240$\pm$0.0040 &   0.0282$\pm$0.0040 &   0.0269$\pm$0.0032\\    \hline
                     
\end{tabular}
\end{table*}

\subsection{Prospective Reconstruction}
\begin{figure*}[!t]
\centerline{\includegraphics[width=16cm]{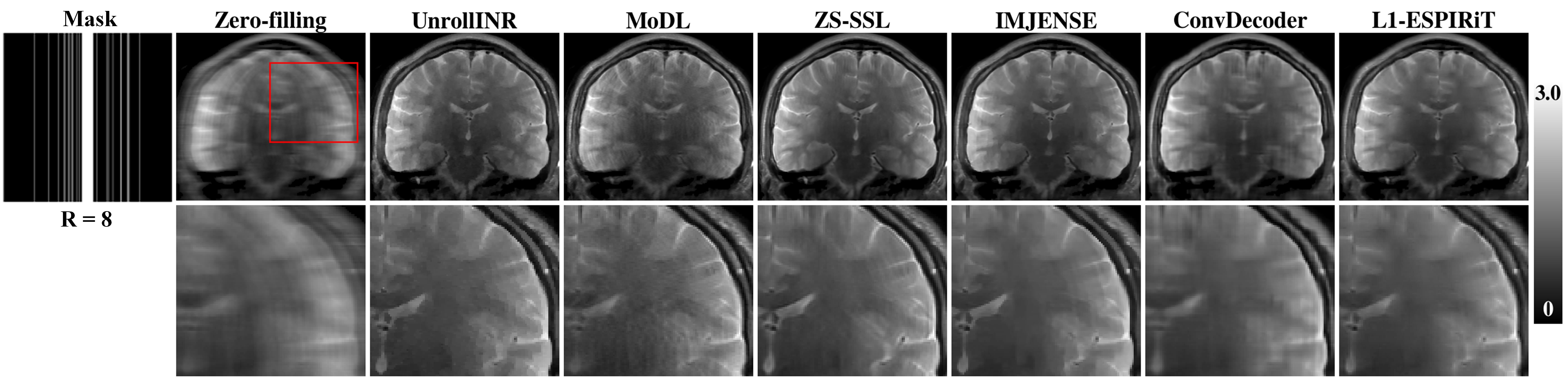}}
\caption{Comparative results of prospective reconstruction for all methods at an acceleration rate R of 8. Locally magnified views of each reconstructed image are provided. The undersampling mask used in the prospective experiment, with an ACS size of 26, is displayed on the leftmost side of the figure.}
\label{fig4}
\end{figure*}

Fig. \ref{fig4} presents the prospective reconstruction results of all comparative methods with an acceleration rate of R = 8 and an actural undersampling rate of 12.5\%. To facilitate comparison, locally magnified views of each reconstructed image are provided. The undersampling mask, which has an ACS size of 26, is displayed on the far left of the figure. As shown in the results, the reconstructed images of L1-ESPIRiT and ConvDecoder appear overly smooth; aliasing artifacts persist in the reconstructions of ZS-SSL and IMJENSE; while the MoDL method exhibits image blurring and noise. In contrast, UnrollINR demonstrates superior performance in artifact suppression and detail preservation, highlighting its precise reconstruction capability.

\subsection{Results of Ablation Study}
\begin{table*}[htbp]
\centering
\caption{Ablation experiment results on the fastMRI knee dataset under different acceleration rates R. The values represent the mean and standard deviation of the quantitative evaluation metrics (PSNR, SSIM, and NRMSE) for 10 slices, with the optimal results highlighted in bold.}
\label{tab3}
\renewcommand{\arraystretch}{1.2}
\begin{tabular}{c|c|c|cccc}
\hline
\textbf{R}&    \textbf{Ablation}&    \textbf{Variant}&      PSNR&      SSIM&   NRMSE&\\ \hline   
\multirow{5}{*}{\textbf{6}} &    \textbf{Baseline}&    \textbf{-}&   \textbf{39.28$\pm$1.83}&     \textbf{0.9590$\pm$0.0090}&  \textbf{0.0111$\pm$0.0021}\\ \cline{2-6}
                   &    \multirow{2}{*}{\textbf{Unrolled network}}&   \textbf{w/o $\mathcal{R}$}&    33.55$\pm$1.22&    0.8903$\pm$0.0195&   0.0212$\pm$0.0029\\    
                   &   &   \textbf{w/o DC}&    35.59$\pm$1.35&    0.9127$\pm$0.0144&   0.0168$\pm$0.0025  \\ \cline{2-6}
                   &    \textbf{Coordinate encoding}&   \textbf{Instant-NGP$\rightarrow$ DINER}&   38.42$\pm$1.76&    0.9518$\pm$0.0104& 0.0122$\pm$0.0023 \\  \cline{2-6}
                   &    \textbf{Loss function}&   \textbf{w/o TV}&    38.94$\pm$1.57&  0.9588$\pm$0.0080&   0.0115$\pm$0.0019 \\   \hline
                   
\multirow{5}{*}{\textbf{10}}&   \textbf{Baseline}&     \textbf{-}&    \textbf{37.58$\pm$1.94}&   \textbf{0.9436$\pm$0.0135}&  \textbf{0.0135$\pm$0.0028}\\   \cline{2-6}                  
                   &   \multirow{2}{*}{\textbf{Unrolled network}}&    \textbf{w/o $\mathcal{R}$}&    31.1$\pm$1.36&    0.8420$\pm$0.0291&   0.0282$\pm$0.0044\\    
                   &   &  \textbf{w/o DC}&    33.34$\pm$1.26&    0.8726$\pm$0.0210&  0.0217$\pm$0.0031 \\ \cline{2-6}  
                   &   \textbf{Coordinate encoding}&    \textbf{Instant-NGP$\rightarrow$ DINER}&    36.35$\pm$1.76&    0.9314$\pm$0.0148&  0.0155$\pm$0.0029\\ \cline{2-6} 
                   &   \textbf{Loss function}&    \textbf{w/o TV}&  37.00$\pm$1.68&   0.9394$\pm$0.0128&   0.0144$\pm$0.0026\\   \hline
\end{tabular}
\end{table*}
Table \ref{tab3} summarizes the quantitative evaluation metrics of reconstruction quality under different ablation study settings at acceleration rates R = 6 and 10. The ablation experiments were conducted on 10 randomly selected slices from the fastMRI knee dataset, with the mean and standard deviation of PSNR, SSIM, and NRMSE reported across these slices.

As shown in Table \ref{tab3}, the complete framework of the proposed UnrollINR consistently achieves the highest reconstruction accuracy. The removal of either the regularization term or the data consistency term leads to a significant decline in reconstruction performance, demonstrating the necessity of these key components in the unrolled network. The impact of the regularization term is particularly notable, underscoring the importance of its design. To evaluate the influence of the coordinate encoding function, the default coordinate encoding method was replaced. The multi-resolution hash encoding Instant-NGP adopted in the baseline outperforms the alternative DINER by better leveraging non-local and local features, and the experimental results confirm the effectiveness of the coordinate encoding approach in improving reconstruction accuracy. In experiments concerning the loss function, the removal of the TV regularization term also resulted in varying degrees of performance degradation. Furthermore, the results indicate that at higher acceleration rates, these key modules play an even more critical role, affirming their necessity for accurate reconstruction.

\section{Discussion}
This study proposes a self-supervised deep unrolled network, UnrollINR, which achieves robust and fast MRI reconstruction by integrating the advantages of deep unrolled structure with INR-based regularization. Validated on both retrospective and prospective datasets containing varying numbers of coil channels, acceleration rates, and ACS sizes, the results demonstrate that even at high acceleration rates, the proposed method effectively suppresses undersampling aliasing artifacts and restores image details. These results highlight the considerable potential of UnrollINR for further accelerating MRI acquisition.

\subsection{Adaptive Tuning of Hyperparameters}
In terms of hyperparameter tuning, this study involves two key hyperparameters: the regularization parameter $\lambda$ in the unrolled framework, and the penalty parameter $\lambda_{s}$ in the loss function. In the proposed UnrollINR model, both $\lambda$ and $\lambda_{s}$ are treated as learnable parameters that can be dynamically updated during training. 
An early stopping strategy is employed, with training terminated after approximately 6000 iterations to prevent the parameters from converging to zero. To determine their initial values, the Ray Tune~\cite{liaw2018tune} automated hyperparameter optimization framework integrated with Optuna is employed to search the parameter space. A randomly selected dataset with fully sampled data serving as reference is used, where Ray Tune optimizes the hyperparameters based on the PSNR between the reconstructed and reference images as the evaluation metric. The hyperparameter search ranges are set as follows: $\lambda$ from 0.01 to 0.10 with a step size of 0.01, and $\lambda_{s}$ from 0.1 to 1.0 with a step size of 0.1.

\begin{figure}[!t]
\centerline{\includegraphics[width=\columnwidth]{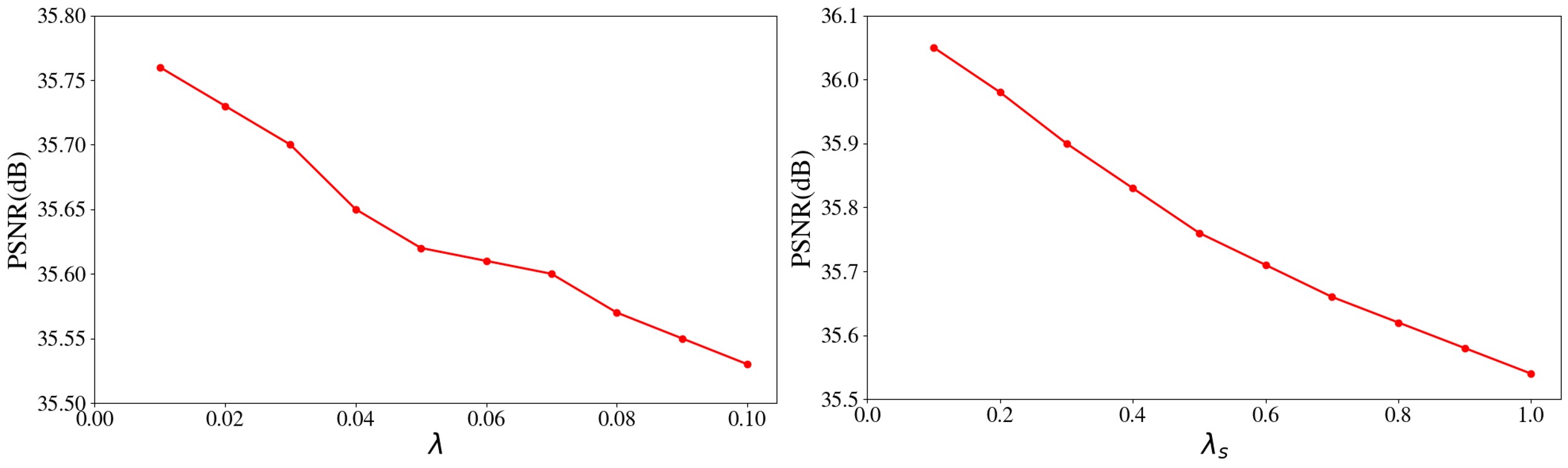}}
\caption{The impact of two key hyperparameters in UnrollINR on PSNR values using the fastMRI knee dataset with an acceleration rate of R = 10.}
\label{fig7}
\end{figure}

To evaluate the sensitivity of the proposed UnrollINR to hyperparameters, experiments were conducted on the fastMRI knee dataset with an acceleration rate R of 10. In each test, while adjusting one hyperparameter, the value of the other hyperparameter remained fixed. The experimental results are documented in Fig. \ref{fig7}. By selecting appropriate initial values for the hyperparameters and incorporating an adaptive tuning strategy, a balance between data consistency and regularization can be achieved, ensuring robust reconstruction performance.

\subsection{Generalizability of UnrollINR to Different Sampling Patterns}
\begin{figure}[!t]
\centerline{\includegraphics[width=\columnwidth]{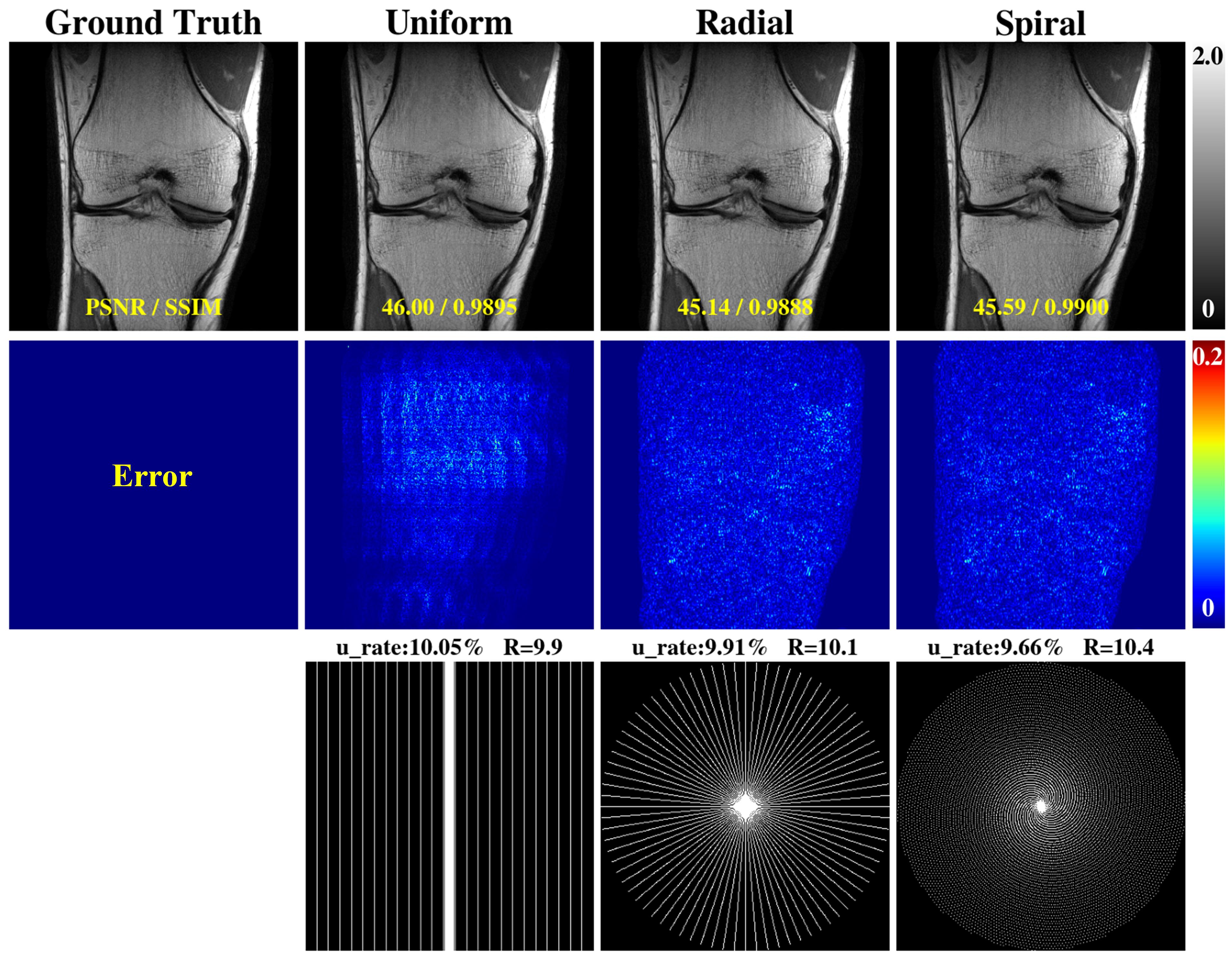}}
\caption{Reconstruction results of UnrollINR on the fastMRI knee dataset under different undersampling patterns. Quantitative metrics (PSNR and SSIM) are indicated at the bottom of each reconstructed image. Absolute error maps between the reconstructed images and the ground truth are also provided. The corresponding undersampling patterns are displayed at the bottom of the figure, with the undersampling rate u\_rate and acceleration rate R annotated for each pattern.}
\label{fig5}
\end{figure}

This study further investigates the adaptability of the proposed UnrollINR to different undersampling patterns. Three representative undersampling patterns were employed: uniform, radial, and spiral. All sampling patterns were set at an acceleration rate of approximately 10. The experimental results, shown in Fig. \ref{fig5}, demonstrate that the proposed UnrollINR maintains consistent and reliable reconstruction performance across various sampling patterns, effectively eliminating the distinct aliasing artifacts introduced by different undersampling patterns. These results indicate that the method is insensitive to undersampling patterns, exhibiting strong robustness and generalization capability.

\subsection{Choice for the Number of CG Iterations}
\begin{figure}[!t]
\centerline{\includegraphics[width=7cm]{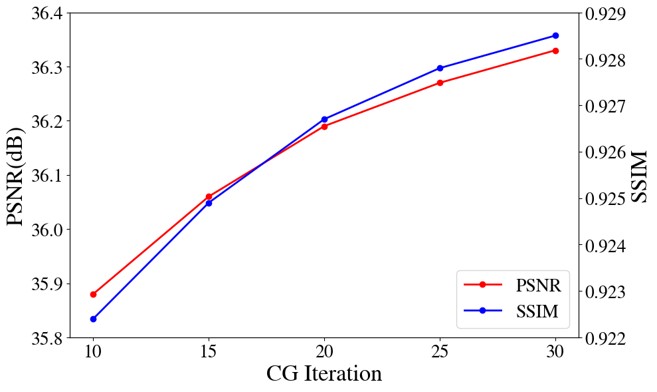}}
\caption{Impact of the number of CG iterations on quantitative metrics, PSNR and SSIM, using the fastMRI knee dataset at an acceleration rate R = 10.}
\label{fig8}
\end{figure}

The number of CG iterations refers to the iteration count within the DC term. To investigate the impact of the CG iteration number on reconstruction performance, five iteration counts ranging from 10 to 30 with an interval of 5 were selected. Experiments were conducted on the fastMRI knee dataset with an acceleration rate of R = 10. As shown in Fig. \ref{fig8}, reconstruction performance improves as the number of CG iterations increases. However, the performance gain diminishes with further increments in the iteration count. Table \ref{tab5} further shows that training time increases substantially with iteration count. Balancing computational cost and reconstruction quality, the optimal CG iteration number was ultimately set to 20.

\begin{table}[htbp]
\centering
\caption{The training and inference times for different number of CG iterations.}
\label{tab5}
\renewcommand{\arraystretch}{1.2}
\begin{tabular}{c|cc}
\hline
\textbf{CG Iteration}&    \textbf{Training (min)}&    \textbf{Inference (s)}    \\ \hline   
10&    11.01&  0.0305\\ 
15&    14.70&  0.0455\\ 
20&    18.53&  0.0587\\ 
25&    22.67&  0.0746\\
30&    26.76&  0.0876\\
\hline  
\end{tabular}
\end{table}





\subsection{Limitation and Future Work}
Despite its promising performance in accelerating MRI reconstruction, the proposed UnrollINR has certain limitations.
As summarized in Table \ref{tab4}, UnrollINR achieves substantially faster training than deep unrolled methods such as ZS-SSL and MoDL, but its overall computational efficiency is still lower than that of lightweight architectures like IMJENSE and ConvDecoder. 
Moreover, both training and inference time are sensitive to the number of CG iterations, particularly for data with larger spatial sizes, which accounts for the differences in runtime observed across datasets. 
Future research could explore alternative INR-based unrolled iterative frameworks to further enhance convergence efficiency. Additionally, advanced acceleration strategies may be considered, such as employing meta-learning to precondition MLP weights for specific signal distributions or leveraging transfer learning to adapt pre-trained models to new reconstruction domains. These extensions have the potential to achieve faster convergence and enhanced reconstruction fidelity when generalizing to unseen data.

\begin{table}[htbp]
\centering
\caption{The training and inference times for all comparison methods.}
\label{tab4}
\renewcommand{\arraystretch}{1.2}
\begin{tabular}{c|c|cc}
\hline
\textbf{Datasets}& \textbf{Methods}&    \textbf{Training (min)}&    \textbf{Inference (s)}    \\ \hline   
\multirow{6}{*}{\makecell{\textbf{FastMRI} \\ \textbf{knee dataset}}}&  UnrollINR (GPU)&    16.40&  0.0585\\ 
&  MoDL (GPU)&    550.14&  0.3759\\ 
&  ZS-SSL (GPU)&    31.95&  0.6923\\ 
&  IMJENSE (GPU)&    0.18&  0.0026\\
&  ConvDecoder (GPU)&    1.57&  0.0039\\
&  L1-ESPIRiT (GPU)&    -&  4.3985\\
\hline
\multirow{6}{*}{\makecell{\textbf{FastMRI} \\ \textbf{brain dataset}}}&  UnrollINR (GPU)&    3.52&  0.0114\\ 
&  MoDL (GPU)&    88.58&  0.3780\\ 
&  ZS-SSL (GPU)&    23.73&  0.6156\\ 
&  IMJENSE (GPU)&   0.09&  0.0013\\
&  ConvDecoder (GPU)&    1.25&  0.0035\\
&  L1-ESPIRiT (GPU)&    -&  2.1830\\ 
\hline  
\end{tabular}
\end{table}

\section{Conclusion}
In this study, a novel zero-shot self-supervised unrolled reconstruction method UnrollINR is proposed, which enables high-quality fast MRI reconstruction without external training data. By effectively integrating unrolled reconstruction architectures with INR, the method significantly enhances MRI reconstruction performance while maintaining model interpretability. Experimental results demonstrate that the proposed method achieves superior reconstruction quality compared to other methods, while maintaining reliable performance even under highly accelerated rates.



\bibliography{UnrollINR_ref}
\bibliographystyle{IEEEtran}

\end{document}